\documentclass[letterpaper, 10 pt, conference]{ieeeconf}  
\IEEEoverridecommandlockouts                           

\usepackage{cite}
\usepackage{amsmath,amssymb,amsfonts}
\usepackage{algorithmic}
\usepackage{graphicx}
\usepackage{textcomp}
\usepackage{xcolor}
\usepackage{float}
\usepackage{caption}
\usepackage{subcaption}
\usepackage{hyperref}
\usepackage{multirow}
\usepackage{amsmath} 
\usepackage{comment}
\usepackage{listings}
\usepackage{xcolor}
\usepackage{algorithm}
\usepackage{algorithmic}
\usepackage{comment}
\usepackage{rotating} 
\usepackage{pdflscape}
\usepackage{cuted}      
\usepackage{capt-of}    
\usepackage{tabularx}
\usepackage{stfloats}
\usepackage[switch]{lineno}

\title{Cross-Dataset Linkage of Brain MRI using Image Similarity Measures

}

\author{
Gaurang Sharma$^{*,1,2}$,
Harri Pölönen$^{1}$,
Juha Pajula$^{1}$,
Jutta Suksi$^{1}$,
and Jussi Tohka$^{2}$\\
for the Alzheimer’s Disease Neuroimaging Initiative$^{\dagger}$%
\thanks{$^{*}$Corresponding author. Email: gaurang.sharma@vtt.fi}%
\thanks{$^{1}$VTT Technical Research Centre of Finland Ltd,
Espoo, Finland.
{\tt\small Email: firstname.lastname@vtt.fi}}%
\thanks{$^{2}$A.I. Virtanen Institute for Molecular Sciences,
University of Eastern Finland, Kuopio, Finland.
{\tt\small Email: firstname.lastname@uef.fi}}%
\thanks{$^{\dagger}$Data used in preparation of this article were obtained from the
Alzheimer's Disease Neuroimaging Initiative (ADNI) database
(\url{https://adni.loni.usc.edu}).
As such, the investigators within the ADNI contributed to the design and implementation of ADNI and/or provided data but did not participate in the analysis or writing of this report. A complete listing of ADNI investigators can be found at:
\url{http://adni.loni.usc.edu/wp-content/uploads/how_to_apply/ADNI_Acknowledgement_List.pdf}}%
}
\begin{document}

\maketitle
\thispagestyle{empty}
\pagestyle{empty}

\begin{abstract}


Head magnetic resonance imaging (MRI) data are routinely collected and shared for research under strict regulatory frameworks that require the removal of direct identifiers prior to data release. However, even after skull stripping, brain parenchyma may retain participant-specific features that enable linkage of scans acquired from the same individual across datasets, posing a potential privacy risk when combined with auxiliary information. Current regulatory approaches typically assess such risks using qualitative notions of reasonableness. Although prior work has suggested that brain MRI can support subject linkage, existing demonstrations have relied on training-based or computationally intensive methods.

Here, we show that reliable linkage of skull-stripped T1-weighted brain MRI is possible using standard preprocessing pipelines followed by direct image similarity computations. Using this simple approach, we achieve near-perfect matching accuracy across datasets acquired at different time points, with varying scanner types, spatial resolutions, and acquisition protocols, and even in the presence of cognitive decline. These experiments simulate realistic scenarios of cross-database matching in large-scale neuroimaging repositories. Our findings highlight a previously underappreciated re-identification risk in shared brain MRI data and provide empirical evidence relevant to the development of informed, forward-looking data-sharing policies in neuroimaging research.

\end{abstract}


\section{INTRODUCTION}

Artificial intelligence (AI) holds the promise of transforming care, enabling earlier diagnoses, tailoring treatments to individual needs, and supporting clinicians at scale. To realize this potential, AI models must be trained with rich, varied datasets drawn from multiple institutions, reflecting the natural complexity of real-world practice. This makes seamless data sharing across sites essential.

However, sharing sensitive health data introduces profound privacy challenges.
When two datasets include records from the same individual even without explicit identifiers, the overlap of implicit features can create hidden connections, enabling linkage of health records across these datasets. This process, known as ``record linkage" \cite{eisinger2025data}, raises significant privacy risks and facilitates various linkage attacks, including record re-identification, attribute inference, and cross-dataset matching \cite{packhauser2022deep, rocher2019estimating, barth2012re, narayanan2008robust, gymrek2013identifying}.

\begin{figure}
    \centering
    \includegraphics[width=1\linewidth]{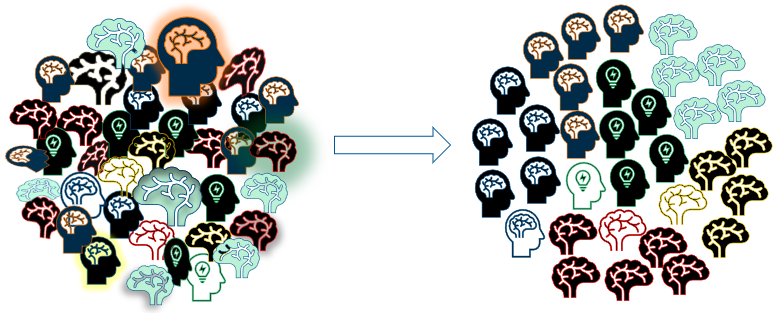}
    \caption{Brain MRI collected across sites, scanners, and timepoints, even after skill-stripping, can retain individual-specific features. When multi-source data (Left) is harmonized, it may inadvertently converge toward participant-specific patterns (Right), effectively linking records and increasing the risk of data samples matching.}
    \label{fig:heroImage}
\end{figure}

Regulatory frameworks relating to privacy and the protection of personal data, such as the European General Data Protection Regulation (GDPR)~\cite{GDPR2016}, the U.S. Health Insurance Portability and Accountability Act (HIPAA) ~\cite{HIPAA1996}, and the Australian Privacy Act~\cite{PrivacyAct1988_C2004A03712}, establish the basis for the practices on the removal of direct identifiers, de-identification, anonymization and pseudonymization. These frameworks also set out minimization principles and risk assessment requirements. However, often the measures required for de-identification or pseudonymization and the risk assessments are dependent on broadly defined terms, such as ``reasonably likely,” ``insignificant risk,” ``reasonable basis,” ``very small,” or ``reasonably identifiable”. Despite further guidance given by the regulatory authorities, the assessment of reasonableness remains ambiguous and dependent on the context.

Magnetic Resonance Imaging (MRI) of the head 
preserves cranial and facial structures that can identify individuals \cite{prior2008facial, mazura2012facial}. 
To improve privacy, researchers often remove or replace information that could identify a person. 
Defacing, completely or partially, removes facial features, retaining structures, such as the skull and skin \cite{bischoff2007technique, schimke2011quickshear, GAO2025111112}, whereas skull-stripping isolates only the brain parenchyma, providing stronger privacy protection \cite{EKE2021100053}. 
However, recent studies have shown that, even after skull stripping, brain parenchyma retains participant-specific anatomical patterns that can link MRIs from the same individual, see Figure~\ref{fig:heroImage}. 
Using supervised learning, Valizadeh et al. and Jäncke et al. demonstrated participant matching from longitudinal T1-weighted scans \cite{valizadeh2018identification, jancke2022identification}. Wachinger et al. proposed BrainPrint for characterizing individual brain morphology and applied the framework for subject identification based on supervised learning \cite{WACHINGER2015232}, which Puglisi et al. extended to DeepBrainPrint, a deep learning framework \cite{puglisi2023deepbrainprint}. Chauvin et al. proposed a keypoint-based similarity based method to match T1w MRI scans from the same participants and demonstrated that scans from close relatives are more similar to each other than scans from unrelated persons \cite{chauvin2020neuroimage}.

Matching a participant’s MRI across databases would allow an adversary to identify other scans from the same participant, associate them with clinical or demographic records, and reconstruct unintended longitudinal health profiles, thereby possibly compromising anonymity and exposing sensitive medical information.
 However, most of the above cited studies rely on learning-based methods that require multiple labeled scans per individual for training, limiting the applicability of similarity estimation when only a single scan is available. Moreover, evaluations are largely restricted to the same-scanner data, short inter-scan intervals, and healthy cohorts, without realistically simulating cross-database scenarios involving different scanners or acquisition protocols. Only Jäncke et al. examined long inter-scan intervals (up to seven years), but their approach still requires matched MRI pairs to identify a third scan \cite{jancke2022identification}.

Thus, we aimed to assess the MRI linkage in an unsupervised manner using simple, open-source, and frequently used techniques.
We simulated MRI matching across databases and focused on the question: Do existing MRI preprocessing methods, when strategically applied, determine whether scans of the same individual, acquired from different sources, at various time points, and under varying protocols, can be linked together? Furthermore, does cognitive status (control, mild cognitive impairment (MCI), or dementia) influence the likelihood of MRI linkage?

Our work addressed these questions within the framework of privacy and data protection regulations, more specifically in respect of the regulatory framework set out by the GDPR. We introduced a structured pipeline that operationalized the concepts used in GDPR with regard to pseudonymized data when assessing, firstly, the likelihood of means of re-identification (``means reasonably likely to be used to identify the natural person”) and, secondly, the residual risk level embedded in the possibility to link it to other information (``insignificant risk”) using widely accessible tools and computationally efficient methods, while remaining aligned with standard neuroimaging practices.

The pipeline incorporates a standard preprocessing-based harmonization workflow for T1-weighted (T1w) MRI, followed by image similarity assessment using a set of commonly used image similarity measures. To ensure robustness, we also evaluated harmonization using standard toolbox-based methods, demonstrating that the results hold across harmonization techniques and conducted an extensive evaluation across diverse datasets spanning multiple imaging protocols, sites, scanners, sessions, and cognitive status. 


To the best of current knowledge, no prior work has evaluated the potential of brain MRI to serve as a linkage mechanism across datasets under regulatory interpretations of reasonableness.

\begin{figure*}[t]
    \centering

    \includegraphics[width=1\linewidth]{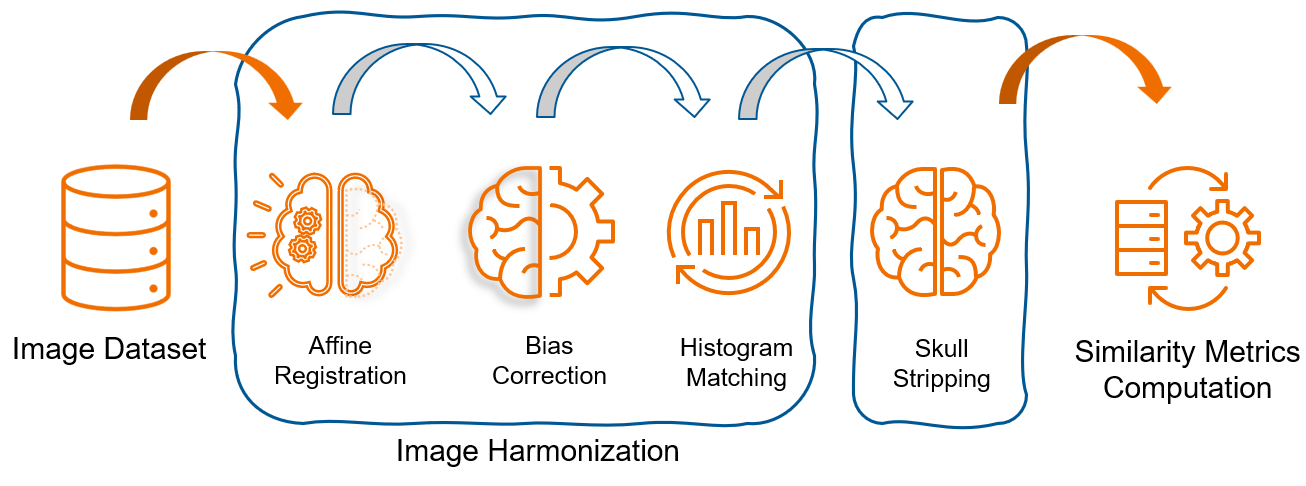}

    \begin{subfigure}[b]{0.19\textwidth}
        \includegraphics[width=\linewidth]{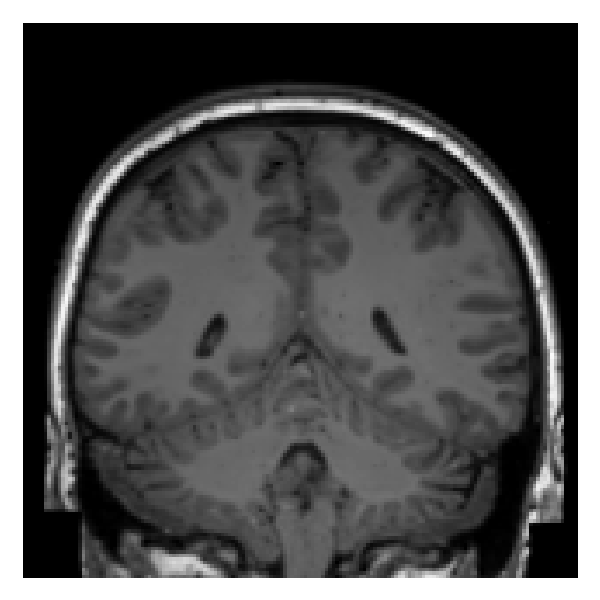}
        \caption{Original}
    \end{subfigure}
    \hfill
    \begin{subfigure}[b]{0.19\textwidth}
        \includegraphics[width=\linewidth]{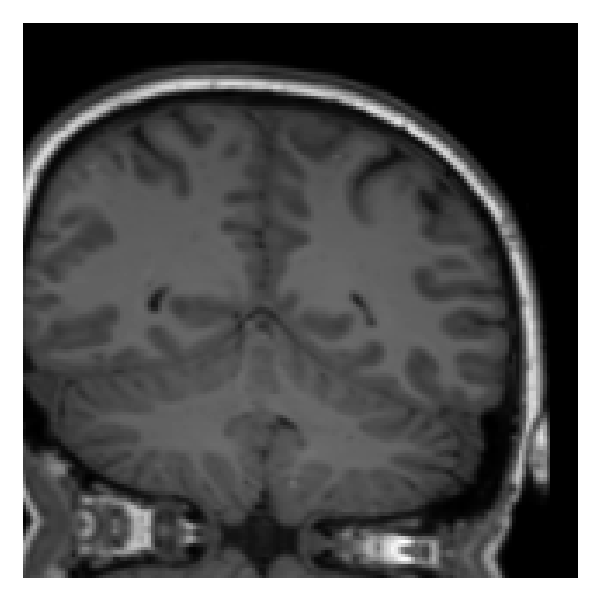}
        \caption{Affine}
    \end{subfigure}
    \hfill
    \begin{subfigure}[b]{0.19\textwidth}
        \includegraphics[width=\linewidth]{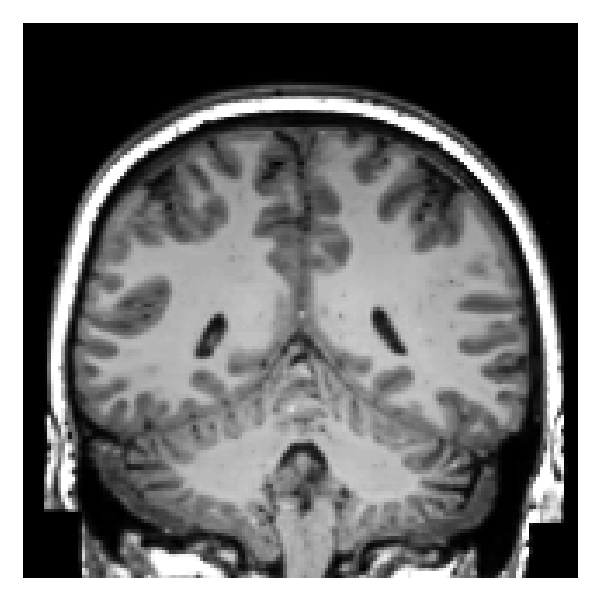}
        \caption{Intensity}
    \end{subfigure}
    \hfill
    \begin{subfigure}[b]{0.19\textwidth}
        \includegraphics[width=\linewidth]{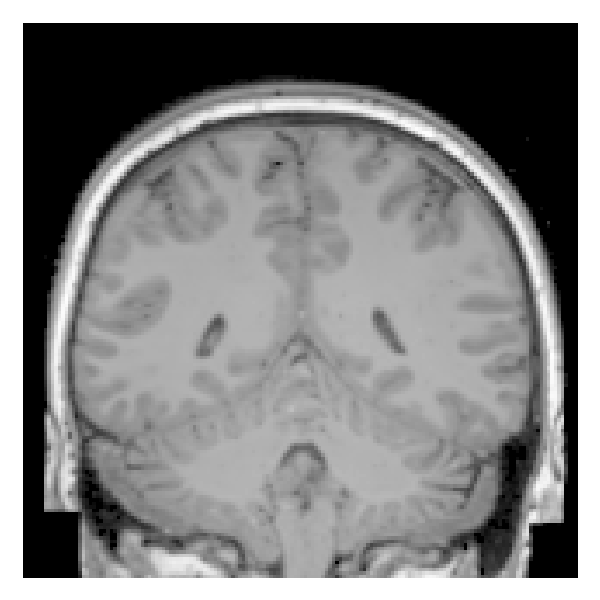}
        \caption{Contrast}
    \end{subfigure}
    \hfill
    \begin{subfigure}[b]{0.19\textwidth}
        \includegraphics[width=\linewidth, height=0.139\textheight]{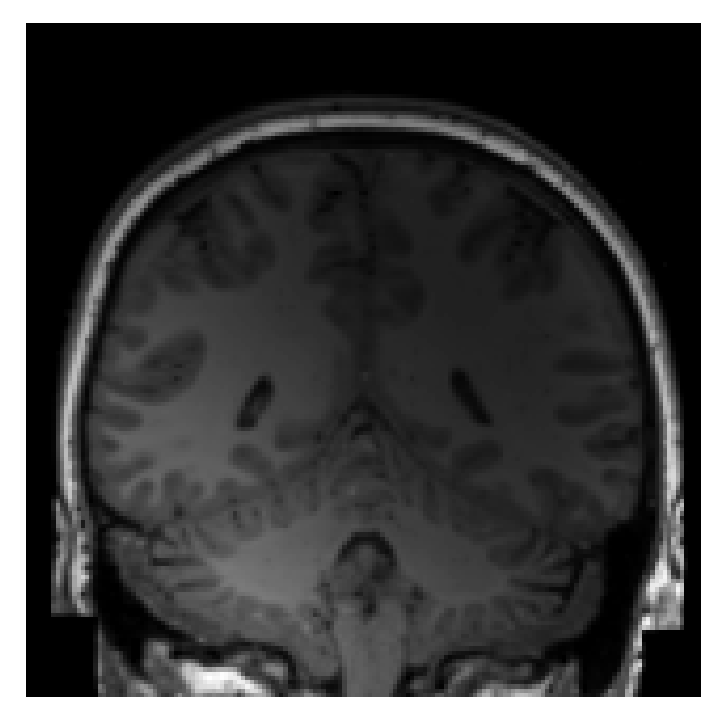}
        \caption{Bias Field}
    \end{subfigure}

    \begin{subfigure}[b]{0.19\textwidth}
        \includegraphics[width=\linewidth]{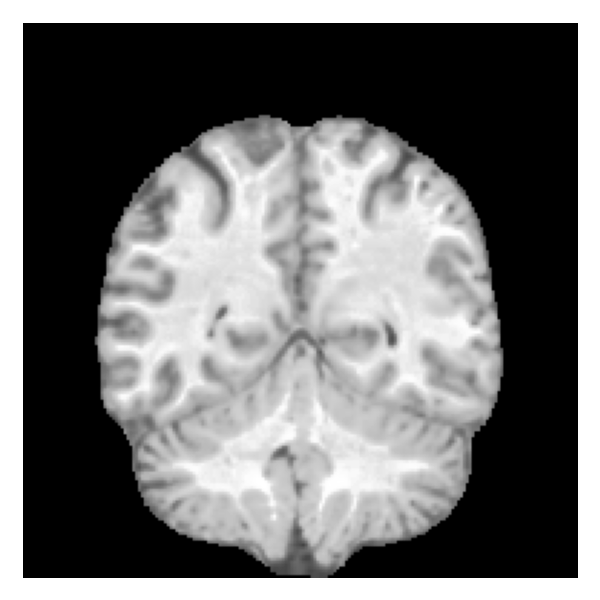}
    \end{subfigure}
    \hfill
    \begin{subfigure}[b]{0.19\textwidth}
        \includegraphics[width=\linewidth]{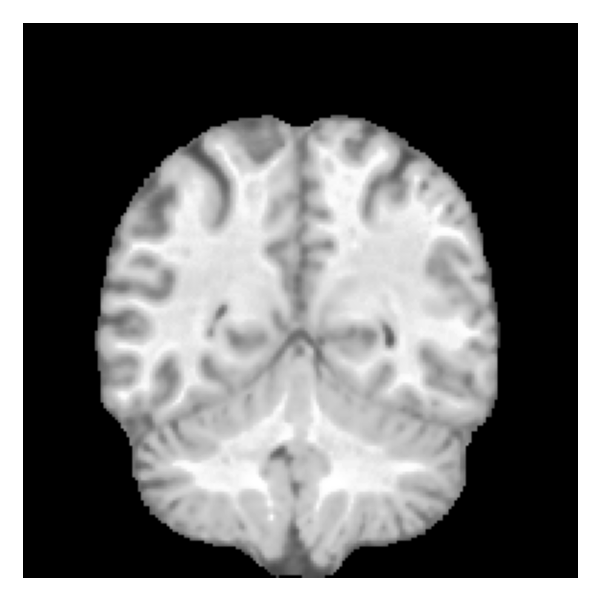}
    \end{subfigure}
    \hfill
    \begin{subfigure}[b]{0.19\textwidth}
        \includegraphics[width=\linewidth]{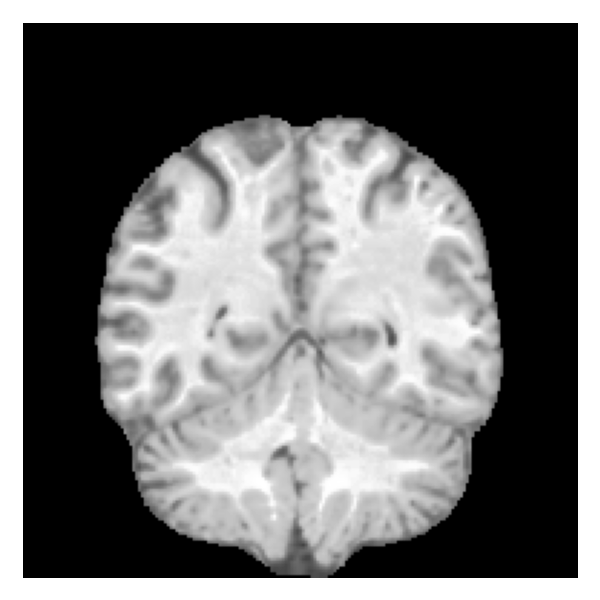}
    \end{subfigure}
    \hfill
    \begin{subfigure}[b]{0.19\textwidth}
        \includegraphics[width=\linewidth]{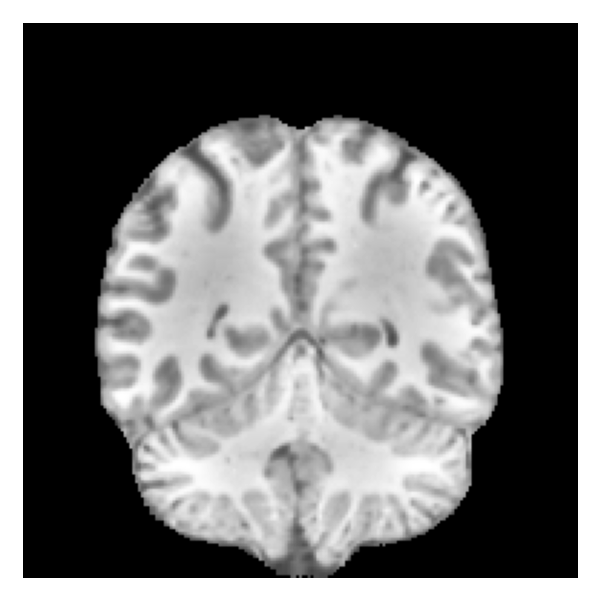}
    \end{subfigure}
    \hfill
    \begin{subfigure}[b]{0.19\textwidth}
        \includegraphics[width=\linewidth]{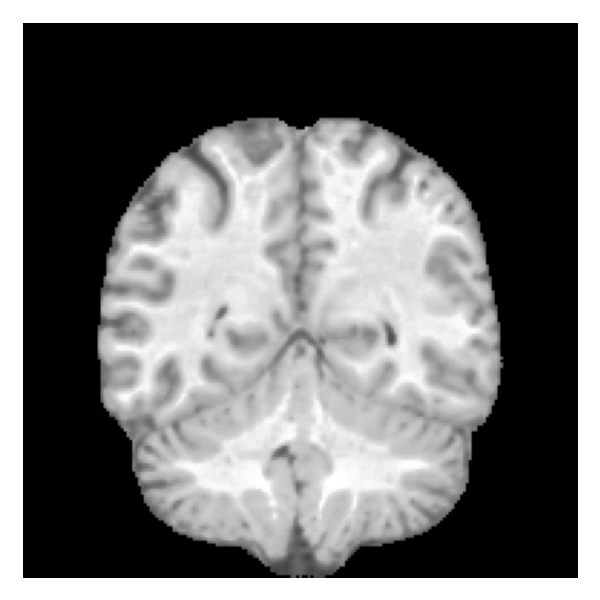}
    \end{subfigure}

    \caption{
    Overview of the pipeline and example transformations of synthetic MRI data from the SLDM dataset.
The top row depicts the full proposed pipeline, which relies on standard MRI processing steps. The middle and bottom rows show an example SLDM image, its transformed variants, and their harmonized outputs, illustrated using a coronal slice at MNI y-coordinate $-15mm$. These examples highlight the pipeline’s ability to standardize both anatomical alignment and intensity distributions. Skull stripping was applied to restrict visualization to the brain.
}\label{fig:combinedPipeline_and_harmonization}
\end{figure*}

\section{Materials and methods}

Our study demonstrates that using standard pre-processing and image similarity measures, it was possible to link a skull-stripped T1-weighted brain MRI from the same person with near-perfect accuracy.
The analysis proceeded in two phases. In the first phase, we tested the approach on two simulated datasets. 
In the second phase, we performed a comprehensive evaluation across five independent publicly available research datasets. The datasets are detailed in  Table~\ref{tab:Study_Data_description}.

This study does not attempt to match individuals across databases or break data silos. All analyses were performed within a single, authorized data environment using pre-existing, study-provided links between MRI scans solely to evaluate methodological claims. No external records were linked, and no attempts were made to re-identify participants, ensuring full compliance with data access and user agreements.

\subsection*{\textnormal{\textbf{Data}}}
Datasets used in pre-evaluation and evaluation phases are summarized in Table \ref{tab:Study_Data_description} and detailed below. 

\subsection*{Simulated Datasets for Pre-Evaluation}
We simulated two datasets for the pre-evaluation phase: Simulated-HCP (SHCP), based on the Human Connectome Project (HCP) Young Adult Dataset \cite{van2013wu}, and Simulated-LDM (SLDM), based on the LDM 100K dataset\cite{pinaya2022brain}. Each simulated dataset had 500 images: 100 original and 400 perturbed. 

The HCP Young Adult dataset includes high-resolution T1-weighted structural MRI scans of 1,112 healthy individuals aged 22 to 37, including twin and non-twin sibling families. These scans were acquired with a customized 3T Siemens Connectom scanner using a 3D MPRAGE sequence at an isotropic voxel size of 0.7 mm$^3$. To simulate the SHCP dataset, 100 images were randomly selected (see \textit{Supplementary Information} for participant IDs). For each image, four transformed variants were generated using MONAI (Full documentation is available at: \url{https://docs.monai.io/en/stable/transforms.html}): an affine-transformed, an intensity-shifted, a contrast-adjusted version (with $\gamma$ randomly sampled from $[0.1, 0.9]$), and a bias-field–augmented version (with polynomial coefficients sampled from predefined uniform ranges, $\mathrm{coeff\_range} = (\mathcal{U}[0.1, 0.2],\ \mathcal{U}[0.1, 0.8])$).
These variations mimicked diverse imaging environments affected by scanner, technician, participant, and environment-specific biases. 


Similarly, we simulated the SLDM dataset (see \textit{Supplementary Information} for IDs), using the LDM 100K synthetic dataset created by Pinaya et al. \cite{pinaya2022brain}. The LDM 100K is a large-scale synthetic brain MRI dataset generated using a latent diffusion framework trained on T1-weighted scans from 31,740 healthy adults aged 44 to 82 years in the UK Biobank. It is publicly available at \cite{ldm100k} and features images with 1 mm\textsuperscript{3} isotropic voxels and dimensions of $160 \times 224 \times 160$ voxels. The selected 100 images were transformed in a manner similar to the SHCP setup, yielding an SLDM dataset of 500 images.

\subsection*{Hormonal Health Study (HHS) Dataset}

The HHS dataset \cite{HHS_Dataset}, \cite{babenko2023dynamic, rizor2024menstrual} includes high-resolution T1w (MPRAGE) MRI scans from 30 healthy, naturally cycling women aged 18 to 29, acquired on a 3T Siemens scanner. All images were defaced, with an isotropic voxel size of $0.93$mm\textsuperscript{3}. Each participant underwent three MRI sessions in a longitudinal study on hormone-driven brain plasticity \cite{rizor2024menstrual}.  

\subsection*{Running Intervention Dataset}

The Running Intervention dataset \cite{ds004937:1.0.0} includes longitudinal defaced MRI scans from 21 healthy young men aged 20 to 31 who participated in a seven-week moderate-intensity running intervention. Each participant underwent four MRI sessions (baseline, pre-, mid-, and post-intervention) using T1w MPRAGE imaging on a 3T Siemens scanner, with isotropic voxel sizes of $1$mm\textsuperscript{3}.

\subsection*{Traveling Human Phantom (THP) Dataset}

The THP dataset at OpenNeuro \cite{ds000206:1.0.0, magnotta2012multicenter} includes high-resolution T1w MRI scans of five healthy adult participants, each scanned at eight sites with Siemens, GE, and Philips scanners. All images were defaced and feature various isotropic voxel sizes, including 1.1mm\textsuperscript{3}, 1.07mm\textsuperscript{3}, and 0.63mm\textsuperscript{3}. Multiple scans per participant and site provide a solid basis for assessing scanner-related variability, with T1w images acquired using standardized 3D sequences such as MPRAGE.

\subsection*{San Diego State University Traveling Participants Dataset (SDSU-TS)}
The SDSU-TS dataset \cite{ds005664:1.1.2} includes defaced MRI scans from nine healthy adults (ages 22--55) collected at San Diego State University (SDSU) with GE Discovery MR750 3T scanner and UC San Diego’s Center for Functional MRI (CFMRI) with Siemens Prisma 3T scanner. Each participant had multiple scans at both sites, typically 7 days apart (1 to 19 days). T1w images were acquired with high-res 3D MPRAGE or FSPGR sequences, voxel sizes of $1$mm\textsuperscript{3} at CFMRI and $0.8$mm\textsuperscript{3} at SDSU.

\subsection*{Alzheimer's Disease Neuroimaging Initiative}

The Alzheimer’s Disease Neuroimaging Initiative (ADNI) (URL: \url{http://adni.loni.usc.edu}) is a public–private partnership launched in 2003 to investigate whether imaging, biomarkers, and clinical assessments can track the progression of mild cognitive impairment (MCI) and early Alzheimer’s disease. For details, visit \url{http://www.adni-info.org}.

ADNI is a longitudinal study structured into five major multi‑year phases. Because ADNI-1 and ADNI-2 employed different MRI acquisition protocols, we treated them as separate datasets. We identified 277 participants who were scannedin both phases (the ADNI-2 protocol is available at: \url{https://adni.loni.usc.edu/wp-content/themes/freshnews-dev-v2/documents/clinical/ADNI-2_Protocol.pdf}); corresponding Participant IDs are provided in the \textit{Supplementary Information}.
For across protocols analysis, we selected each participant’s first ADNI-2 scan and paired it with an ADNI-1 scan chosen based on cognitive status at the time of ADNI-1 MRI acquisition. For cognitively normal (CN) participants, we selected an ADNI‑1 scan acquired within three years of the ADNI‑2 scan; for MCI and dementia participants, we chose the closest scan within the same time window. If no scan satisfied this criterion, the nearest available scan was used. 
Table~\ref{tab:participantCount} shows participant counts by time gap and cognitive status.
The final dataset comprised 454 scans from 227 participants. Of these, 16 progressed from CN to MCI, 19 from MCI to dementia, none converted directly from CN to AD, and 2 reverted from MCI to CN.

\begin{table}[htbp]
\centering
\begin{tabularx}{\linewidth}{|X|c|c|c|c|}
\hline
\textbf{Gap (months)} & \textbf{Controls} & \textbf{MCI} & \textbf{Dementia} & \textbf{Total} \\ \hline
12  & 18 & 8  & 10 & 36  \\ \hline
24  & 51 & 41 & 34 & 126 \\ \hline
36  & 22 & 21 & 13 & 56  \\ \hline
42  & 0  & 0  & 1  & 1   \\ \hline
48  & 1  & 4  & 1  & 6   \\ \hline
60  & 1  & 0  & 0  & 1   \\ \hline
108 & 1  & 0  & 0  & 1   \\ \hline
\end{tabularx}
\caption{Acquisition gap between MRI from ADNI1 and ADNI2, along with the participant count distribution by diagnosis at the ADNI-1 imaging session.}
\label{tab:participantCount}
\end{table}

\subsection*{\textnormal{\textbf{Harmonization of MRI datasets}}}

T1-weighted MRI scans underwent harmonization through spatial and intensity standardization using two approaches: (1) a custom pipeline applying widely used open-source software tools (described in detail below)  and (2) processing with the CAT12 toolbox (~\cite{gaser2024cat}, \url{https://neuro-jena.github.io/cat/}, version~9.0). We used different approaches to demonstrate that the results remain consistent across harmonization techniques. The images from the ADNI dataset were processed using CAT12 with the default settings, and all the other images were processed with the custom pipeline. 


The custom pipeline consisted of the following steps (Figure~\ref{fig:combinedPipeline_and_harmonization}):

\begin{enumerate}
\item \textbf{Affine Registration:} Alignment to the MNI ICBM 152 T1 atlas \url{https://www.bic.mni.mcgill.ca/ServicesAtlases/ICBM152NLin2009}  \cite{fonov2011unbiased} using ANTs affine registration (\url{https://antspy.readthedocs.io/en/stable/} with code at \url{https://github.com/ANTsX/ANTsPy}), resizing to 1mm$^3$ voxels \cite{avants2011reproducible}.
\item \textbf{Intensity Harmonization:} Normalization using Z-score method and bias-corrected with SimpleITK 2.4.0's \texttt{N4BiasFieldCorrectionImageFilter}. Histogram matching using the standard atlas, implemented with \texttt{skimage.exposure}'s \texttt{match\_histograms}. Bias correction and histogram matching used brain masks generated using \texttt{deepbet}\cite{fisch2024deepbet}, available at \url{https://github.com/wwu-mmll/deepbet}.
\item \textbf{Skull Stripping:} After harmonization, brain masks were regenerated (using \texttt{deepbet}) to ensure accurate anatomical masking. All images then underwent skull stripping,  and similarity measures were computed exclusively within the brain region to exclude non-brain signals.
\end{enumerate}

ADNI data was first converted to the harmonized Brain Imaging Data Structure (BIDS) data format with Clinica \cite{samper2018reproducible,routier2021clinica} and then was harmonized using CAT12. The harmonization included non-uniformity correction \cite{ashburner2005unified}, adaptive non-local means denoising \cite{manjon2010adaptive}, and registration to a stereotactic space \cite{ashburner2011diffeomorphic}. Here, we used non-linear registration as it is the standard option in the CAT12 pipeline. 


\subsection*{\textnormal{\textbf{Similarity assessment on harmonized images}}}


All unique image pairs \((X, Y)\) were then generated, excluding self-pairings. For each pair, voxel-wise comparison was performed between the two images.

During a pre-evaluation stage, similarity \(S(X, Y)\) was quantified for each pair using 11 measures (see Table~\ref{tab:similarity_metrics} for overview and Appendix for details) that captured both structural and intensity-based characteristics. All metrics were rescaled so that higher values consistently indicate greater similarity.
Based on pre-evaluation performance and computational cost, four measures were selected for the final similarity analysis.

\subsection*{\textnormal{\textbf{Thresholding similarity measures for participant matching}}}

In the pre-evaluation datasets, we considered each original image and its four transformed variants as images from the same participant (forming intra-participant image pairs). In the evaluation datasets, the images were matched based on participant IDs, with any two images from the same participant forming an intra-participant image pair. An image pair with two images from different participants is termed an inter-participant image pair. Note that this ground truth labeling was not used in our similarity-based matching pipeline, but the pipeline is unsupervised.   

We predicted pairwise similarity labels using a clustering-based thresholding
approach. For each similarity measure, we grouped similarity scores into two
clusters using a threshold $\tau$. Given a similarity score $S(X,Y)$, the
predicted label $\hat{y}$ was assigned as
\[
\hat{y} =
\begin{cases}
\text{intra-participant}, & \text{if } S(X,Y) \geq \tau, \\
\text{inter-participant}, & \text{otherwise}.
\end{cases}
\]

To estimate a robust threshold $\tau$, we compared Gaussian Mixture Models
(GMM), Kernel Density Estimation (KDE) clustering, and Otsu’s method. KDE provided the
clearest separation between intra- and inter-participant similarity
distributions and was therefore selected. In short, our KDE-based clustering procedure was as follows: 
We estimated smooth probability densities using Gaussian kernel density estimation implemented in SciPy~\url{https://docs.scipy.org/doc/scipy/tutorial/stats/kernel_density_estimation.html} and identified the global minimum between the two dominant modes using a peak–valley detection procedure based on SciPy’s peak-finding algorithm~\url{https://docs.scipy.org/doc/scipy/reference/generated/scipy.signal.find_peaks.html}. To improve robustness, we removed the top and bottom $2.5\%$ of similarity scores prior to density estimation, except for HHS, which has very few imaging instances, and ADNI, a cross-dataset study with fewer image pairs. We used a bandwidth parameter of $0.25$ and $2000$ evaluation grid
points for all measures.

\subsection*{\textnormal{\textbf{Quantitative performance measures}}}

We quantified performance in the pre-evaluation phase by measuring the overlap between ground-truth similarity distributions as $\sum_{b=1}^{B} \min(P_b, Q_b)$, where $P_b$ and $Q_b$ denote the normalized frequencies in the $b$-th bin of the similarity histograms ($B=100$) for intra- and inter-participant pairs, respectively~\cite{swain1991color}. Overall performance was assessed by comparing predicted labels $\hat{y}_{ij}$ (and similarity scores $s_{ij}$) with ground-truth labels $y_{ij}$ using the area under the ROC curve (AUC), sensitivity, and specificity. where $i$ and $j$ index the two images (or participants) forming each similarity pair.
AUC was computed using the similarity scores of each image pair and the corresponding ground-truth intra- and inter-participant labels. Sensitivity and specificity were calculated using predicted labels obtained by thresholding the similarity measures relative to the ground truth.

\section{Results}



\subsection*{\textnormal{\textbf{Pre-Evaluation Phase: Demonstrating Feasibility of Accurate MRI matching on Simulated MRI Datasets}}}

The pre-evaluation phase used two simulated datasets, SHCP and SLDM, each having 500 images: 100 original and 400 perturbed. Images were first harmonized (see Figure~\ref{fig:combinedPipeline_and_harmonization}) and then paired for voxel-wise similarity comparison. If the similarity $S(X,Y)$ was larger than a threshold $\tau$ then we predicted that images $X$ and $Y$ were from the same participant (intra-participant), and if $S(X,Y)$ was smaller than $ \tau$, then we predicted that the images were from different participants (inter-participant). The ground-truth, i.e., the knowledge of whether the two images truly were from the same participant, was only used for evaluation.

\begin{figure*}[]
    \centering
        \includegraphics[width=\textwidth]{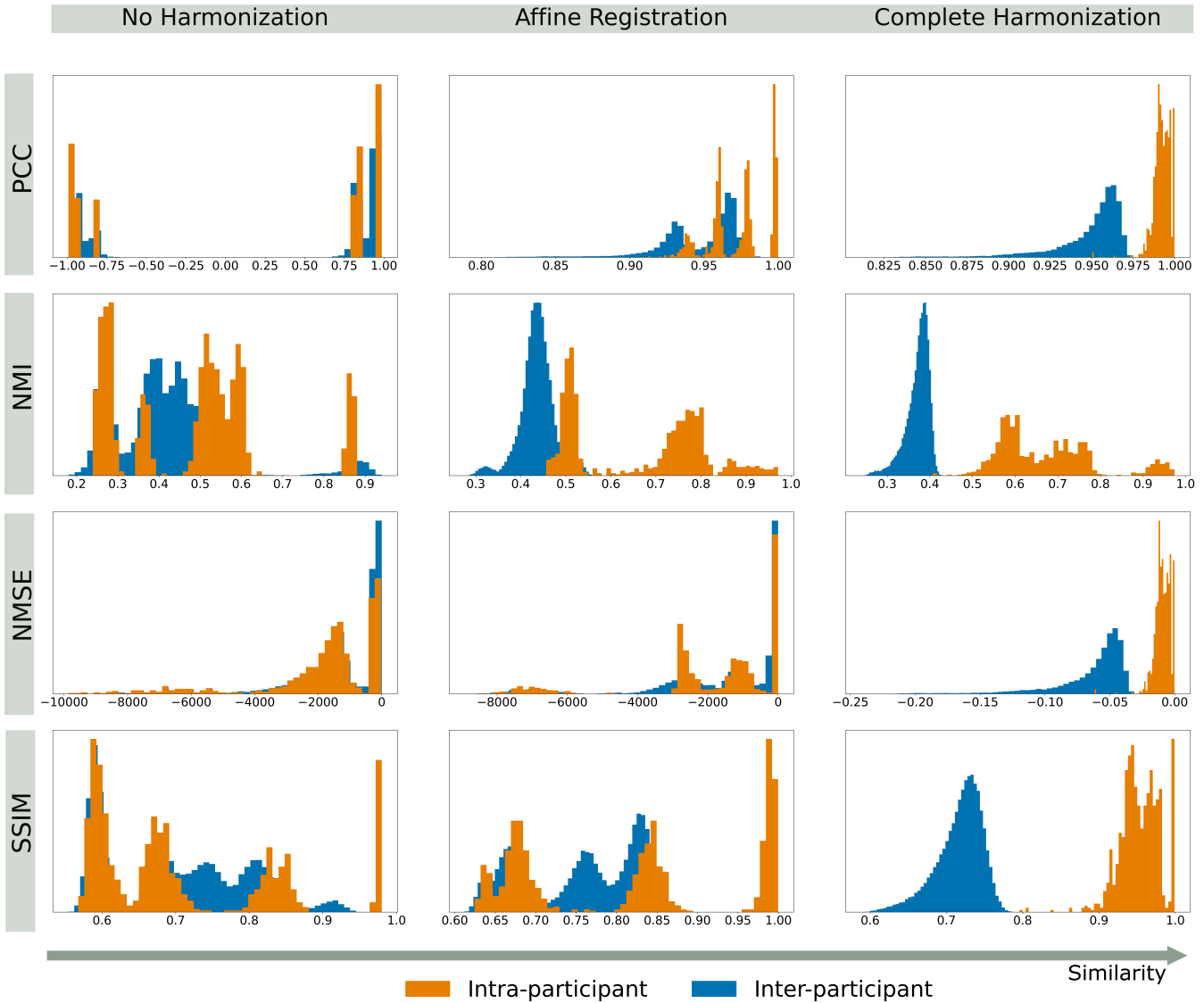}
    \caption{Pre-evaluation on SHCP dataset demonstrates that both intensity and anatomical harmonization were needed to separate inter-participant scan-pairs from intra-participant scan pairs based on similarity measures. Intra and inter-participant labels are based on ground-truth, but are not used in the computation of similarity measures.
    }
    \label{fig:fig_hcp_metrics}
\end{figure*}

Figure~\ref{fig:fig_hcp_metrics} illustrates the distributions of four similarity measures in the SHCP dataset across three harmonization stages—no harmonization, affine registration (anatomical alignment), and full harmonization incorporating both anatomical and intensity alignment. As expected, reliable MRI matching based on similarity scores required both anatomical and intensity harmonization.

\begin{figure*}[]
    \centering
        \includegraphics[width=\textwidth]{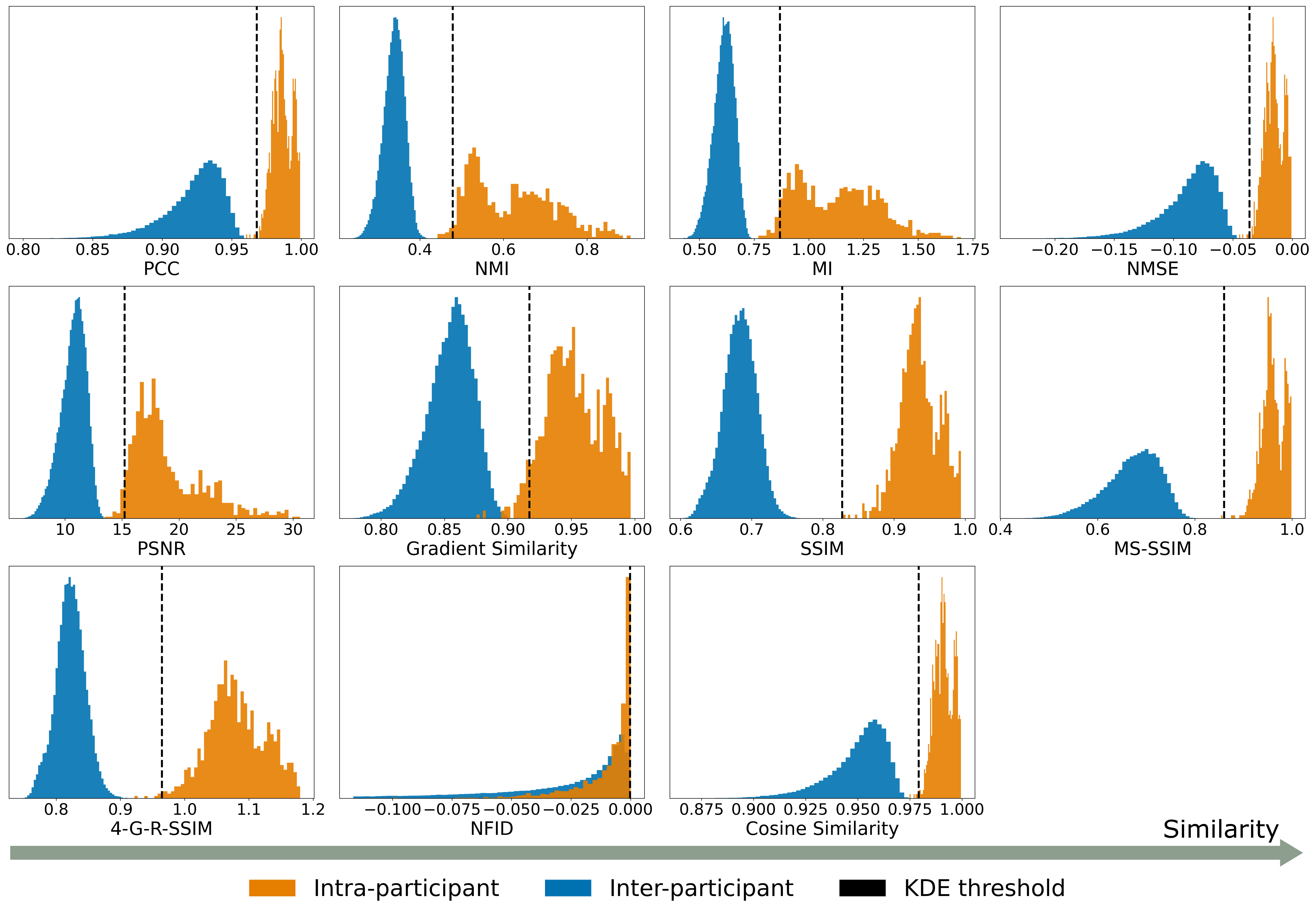}
    
    \caption{Pre-evaluation on SLDM, after harmonization, demonstrates that all measures except NFID clearly distinguish between inter- and intra-participant clusters. Unsupervised thresholds predicted labels for each image pair, yielding an AUC of 1.000 and with a high sensitivity of 0.978--0.999, except for NFID, which had an AUC of 0.82 and a sensitivity of 0.121. To visualize, we removed high-NFID outliers (10{,}573 negative FID values \(< -0.5006)\) using the interquartile range (IQR) method. Intra and inter-participant labels are based on ground-truth, but not used in the computation of similarity measures or in thresholding.
    }
    \label{fig:fig_synth_metrics}
\end{figure*}

Figure~\ref{fig:fig_synth_metrics} presents the post-harmonization distributions of similarity measures for the SLDM dataset, with their quantitative analyses provided in \textit{Supplementary Information Table 1} and the corresponding SHCP results and analyses presented in \textit{Supplementary Information Figure 1 and Table 2}, respectively. 
On the SHCP dataset, Structural Similarity Index (SSIM) and Four-Component Gradient-Regularized SSIM (4‑G‑R‑SSIM) showed the strongest separation between intra-participant and inter-participant pairs with perfect (1.000) AUC values, followed by Normalized Mutual Information (NMI), Multi-Scale-SSIM (MS-SSIM), Gradient Similarity (GradSim), and Pearson’s Correlation Coefficient (PCC), which still exhibited excellent separation and near-perfect performance on AUC, sensitivity, and specificity. Negative Fr\'echet Inception Distance (NFID) showed the least separation. 
Similarly, on the SLDM dataset, most measures showed no overlap with near-perfect AUC, specificity, and sensitivity; only PCC, Peak Signal-to-Noise Ratio (PSNR), Cosine Similarity (CosSim), and Negative Mean Squared Error (NMSE) showed near-zero overlap, with AUC above 0.999. NFID again showed the weakest performance, with a maximum overlap of 0.534 and a minimum AUC of 0.749. 

Harmonizing a single image, using our pipeline, on a CPU with Intel Xeon E5-2698 v4 processor (20~cores, 40~threads) took 87 seconds—10 for registration and 77 for bias correction and histogram matching. Computing all 11 similarity measures required 6 seconds. With 15 concurrent processes, performance improved: for 15 images (105 pairs), harmonization and similarity measure computation required 590 and 410 seconds, respectively.
Table~\ref{tab:similarity_metrics} shows computation times for each metric on CPU and GPU. Details on harmonization timings of the CAT12 are provided in~\cite{gaser2024cat}.

Comparison of image quality measures showed that SSIM, PCC, NMI, and GradSim delivered the best balance of performance and speed, making them our choices for the evaluation phase.

\begin{table*}[htbp]
\centering
\caption{Dataset summaries: Simulated pre-evaluation datasets verified the unsupervised pipeline functionality. Evaluation datasets were used to assess robustness against real-world variability across sites, scanners, protocols, and changes in cognitive performance. ${\ast}$ denotes that the participants are randomly sampled from the original dataset.}
\renewcommand{\arraystretch}{1.2}
\begin{tabularx}{\linewidth}{|X|X|l|l|l|X|}
\hline
  \textbf{Name} &
\textbf{Phase} &
  \textbf{Participants} &
  \textbf{Imaging Instances} &
  \textbf{Age range} &
  \textbf{Scanners Used} \\ \hline
Simulated HCP Young Adult dataset (SHCP) &
Pre-Evaluation &
  $100^{\ast}$ &
  500 &
  22 to 37 years &
  3T Siemens 
  \\ \hline
  Simulated LDM 100K Dataset  (SLDM) &
Pre-Evaluation &
  $100^{\ast}$ &
  500 &
  -- &
  -- %
  \\ \hline
  Hormonal Health Study (HHS) &
Evaluation (Longitudinal) &
  30 &
  66 &
  18 to 29 years &
  Siemens 3T Prisma 
  \\ \hline
  Running intervention Dataset &
Evaluation (Longitudinal) &
  21 &
  84 &
  20 to 31 years &
  3T Siemens 
  \\ \hline
  Traveling Human Phantom &
Evaluation (Multi-site and multi-scanner) &
  5 &
  67 &
   &
  Siemens 3T TIM Trio and Philips 3T Achieva 
  \\ \hline
  San Diego State University Traveling Participants Dataset (SDSU-TS) &
Evaluation (Multi-site and multi-scanner) &
  9 &
  44 &
  22 to 55 years &
  Siemens Prisma 3T and GE Discovery MR750 3T 
  \\ \hline
  Alzheimer’s Disease Neuroimaging Initiative (ADNI) &
  Evaluation (multi-protocol, varying cognitive status, multi-scanner, and multi-site) &
  227 &
  454 &
  55 to 90 years &
  Siemens (TrioTim, Skyra, Prisma), GE (Discovery 750, Signa HDxt, etc.), and Philips (Achieva). 
  ADNI-1 relied primarily on 1.5T MRI, whereas ADNI-2 shifted to 3 T MRI.
  \\ \hline

\end{tabularx}%

\label{tab:Study_Data_description}
\end{table*}

\subsection*{\textnormal{\textbf{Perfect accuracy in linking MRIs within longitudinal samples}}}
Table \ref{tab:evaluation_results} details the performance of selected similarity measures on the datasets used in the evaluation phase.
Evaluation on two longitudinal datasets, the Hormonal Health Study (30 naturally cycling women) and a running intervention dataset (21 healthy young men), achieved perfect scores (AUC, sensitivity, and specificity of $1.000$) in classifying MRI pairs from the same participant. This demonstrates a clear divergence (refer to \textit{Supplementary Information Figures 2 and 3}) between the inter-participant and intra-participant histograms (measure values) across the selected measures.

\subsection*{\textnormal{\textbf{Cross-site MRI scans are linkable despite scanner variability}}}

The Traveling Human Phantom dataset comprises images from five adult participants scanned at eight imaging sites across various timepoints using Siemens, GE, and Philips 3T scanners. Post-harmonization, we achieved an AUC above $0.961$ on all four measures, whereas PCC, despite exhibiting expected inter- and intra- participant distributions, failed to enable participant matching because of ineffective thresholding caused by image artifacts, leading to outliers and insufficient sample size for clustering  (see Table \ref{tab:evaluation_results} and \textit{Supplementary Information Figure 4}).

For the SDSU-TS dataset, which includes images from nine adult participants scanned at two sites using GE Discovery and Siemens Prisma 3T scanners, we achieved near-perfect performance across all measures. The AUC was above $0.986$ for all measures (See Table \ref{tab:evaluation_results} for details and \textit{Supplementary Information Figure 5} for visualizations).

Both datasets contained images with different isotropic voxel sizes, yet the results showed no deviation, demonstrating that standard pre-processing effectively mitigates voxel-size differences when linking MRI records.

\begin{table*}[htbp]
\centering
\caption{Measures for similarity assessment.}
\renewcommand{\arraystretch}{1.2}
\begin{tabularx}{\textwidth}{|p{0.2\textwidth}|X|p{0.14\textwidth}|p{0.14\textwidth}|}
\hline
\textbf{Measures} & \textbf{Description} & \textbf{Computation Time, GPU (in seconds)} & \textbf{Computation Time, CPU (in seconds)} \\
\hline
Mutual Information (MI) & Mutual information between intensity distributions (utilizing joint and marginal probabilities) & 0.020817 & 0.323622\\
\hline
Normalized Mutual Information (NMI) & Normalized mutual information for scale invariance & 0.020877 & 0.324122\\
\hline
Negative Mean Squared Error (NMSE) & Mean squared difference between voxel intensities & 0.000054 & 0.002738\\
\hline
Peak Signal-to-Noise Ratio (PSNR) & Logarithmic measure of peak signal relative to background noise (derived from MSE) & 0.000075 & 0.000167\\
\hline
Pearson Correlation Coefficient (PCC) & Voxel-wise linear correlation between image intensities & 0.000356 & 0.009296\\
\hline
Cosine Similarity (CosSim) & Angular similarity between vectorized image intensities & 0.000911 & 0.005488\\
\hline
Gradient Similarity (GradSim) & Similarity in gradient fields (edge alignment) & 0.014115 & 0.223849\\
\hline
Structural Similarity Index (SSIM) & Visual similarity through local luminance, contrast, and structure comparison & 0.015501 & 0.450654\\
\hline
Multi-Scale Structural Similarity (MS-SSIM) & Multi-scale version of SSIM & 0.020135 & 0.452544\\
\hline
4-G-R-SSIM (Four-Component Gradient-Regularized SSIM) & Rotation-invariant version of SSIM for fine-grained evaluation & 0.271282 & 1.651035\\
\hline
Negative Fréchet Inception Distance (NFID) & Distance between feature distributions (Inception/Med3D) & 2.276406 & 2.595832\\
\hline
\end{tabularx}

\label{tab:similarity_metrics}
\end{table*}

\begin{table*}[htbp]
\centering
\caption{Evaluation across five diverse public datasets, including longitudinal, multi-protocol, and cross-site scenarios, shows that similarity scores after unsupervised thresholding perform strongly. The ${\ast}$ denotes failed cases due to poor image quality at two sites in the Traveling Human Phantom dataset, resulting in failed threshold estimation.}
\renewcommand{\arraystretch}{1.2}
\begin{tabularx}{\linewidth}{|p{5cm}|X|X|X|X|}
\hline
\textbf{Datasets} &
  \textbf{Measures} &
  \textbf{AUC} &
  \textbf{Sensitivity} &
  \textbf{Specificity} \\ \hline

\multirow{4}{*}{\textbf{ADNI}}
  & SSIM    & 1.000 & 1.000 & 1.000 \\ \cline{2-5}
  & NMI     & 1.000 & 0.996 & 1.000 \\ \cline{2-5}
  & GradSim & 1.000 & 0.991 & 1.000 \\ \cline{2-5}
  & PCC     & 1.000 & 1.000 & 1.000 \\ \hline

\multirow{4}{*}{\textbf{Running Intervention}}
  & SSIM    & 1.000 & 1.000 & 1.000 \\ \cline{2-5}
  & NMI     & 1.000 & 1.000 & 1.000 \\ \cline{2-5}
  & GradSim & 1.000 & 1.000 & 1.000 \\ \cline{2-5}
  & PCC     & 1.000 & 1.000 & 1.000 \\ \hline

\multirow{4}{*}{\textbf{Traveling Human Phantom}}
  & SSIM    & 0.963 & 0.899 & 1.000 \\ \cline{2-5}
  & NMI     & 0.963 & 0.901 & 1.000 \\ \cline{2-5}
  & GradSim & 0.962 & 0.858 & 1.000 \\ \cline{2-5}
  & PCC     & 0.961 & 0.997 & $0.072^{\ast}$ \\ \hline

\multirow{4}{*}{\textbf{Hormonal Health Study}}
  & SSIM    & 1.000 & 1.000 & 1.000 \\ \cline{2-5}
  & NMI     & 1.000 & 1.000 & 1.000 \\ \cline{2-5}
  & GradSim & 1.000 & 1.000 & 1.000 \\ \cline{2-5}
  & PCC     & 1.000 & 1.000 & 1.000 \\ \hline

\multirow{4}{*}{\textbf{SDSU-TS}}
  & SSIM    & 0.999 & 0.979 & 1.000 \\ \cline{2-5}
  & NMI     & 0.994 & 0.979 & 1.000 \\ \cline{2-5}
  & GradSim & 0.994 & 0.957 & 1.000 \\ \cline{2-5}
  & PCC     & 0.986 & 0.979 & 1.000 \\ \hline

\end{tabularx}

\label{tab:evaluation_results}
\end{table*}

\subsection*{\textnormal{\textbf{MRI matching remains highly probable regardless of potential cognitive decline, scanner type, or the time interval between MRI acquisitions}}}

We used the ADNI dataset to assess how scanner differences, cognitive decline, and long inter-scan intervals jointly affect participant matching. We analyzed 227 scan pairs from the same individuals across ADNI‑1 (1.5T) and ADNI‑2 (3.0T), during which many participants also showed cognitive decline due to ADNI’s dementia‑focused design. To demonstrate the robustness of the similarity score based matching across different harmonization pipelines, we pre-processed ADNI data through CAT12 pipeline \cite{gaser2024cat}.

\begin{figure*}[]
    \centering
    \includegraphics[width=\textwidth]{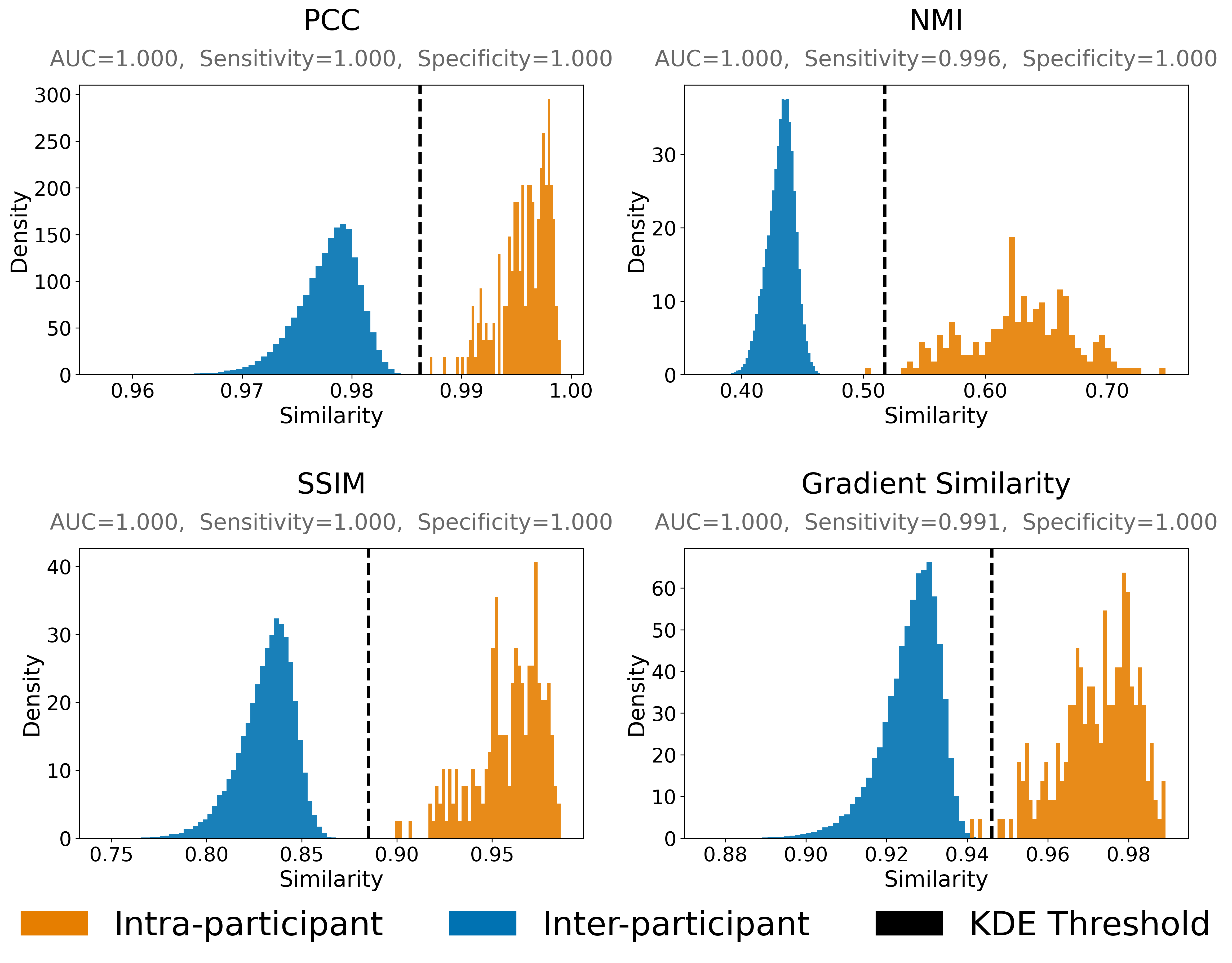}
    \caption{Evaluation on the multi-protocol study demonstrates robust MRI matching across ADNI protocols. For each query image from ADNI1, the corresponding image in ADNI2 was correctly matched, achieving an AUC and specificity of 1.00. Sensitivity exceeded 0.99 overall and reached 1.00 when using SSIM and PCC. Performance remained stable across different imaging protocols, despite potential cognitive decline, and acquisition intervals. Intra and inter-participant labels are based on ground-truth, but not used in the computation of similarity measures nor in thresholding.}
    \label{fig:ADNIresults}
\end{figure*}

As demonstrated in Figure~\ref{fig:ADNIresults} 
SSIM and PCC achieved perfect performance, with AUC, sensitivity, and specificity all equal to $1.000$. The remaining measures also attained an AUC and specificity of $1.000$, with sensitivities exceeding $0.990$. The perfect classification accuracy with SSIM and PCC across all cognitive statuses (normal, mild cognitive impairment, and dementia) demonstrates the pipeline’s robustness to structural brain changes over time. Within this cohort, 16 participants progressed from CN to MCI and 19 advanced from MCI to AD. Most participants’ MRIs were acquired within a 36-month interval, yet even participants with inter-scan intervals of 42, 48, 60, and 108 months were matched across protocols. \textit{Supplementary Information Figure 6} visualizes the structural change in two participants' MRIs over time.

\subsection*{\textnormal{\textbf{Cross-dataset consistency of estimated thresholds}}}

Thresholds obtained by unsupervised clustering were consistent across databases Table~\ref{tab:thresholding_values}) and they showed an excellent balance of sensitivity and specificity~\ref{tab:evaluation_results}. However, when fewer image pairs were available in the Traveling Human Phantom dataset, the resulting thresholds were not optimal in terms of sensitivity/specificity balance, and this was visible in the threshold values that diverged from the average threshold across datasets. This was particularly apparent with PCC. We re-evaluated Traveling Human Phantom and PCC similarity using the mean pre-evaluation threshold. The pre-evaluation threshold provided a favorable trade-off. For PCC, the mean pre-evaluation threshold (0.974) outperformed the KDE-based threshold (0.781), yielding a sensitivity of 0.873, and a specificity of 1.000. This indicates pre-evaluation thresholds, based on synthetic data, could be a viable solution for MRI matching in small datasets.

\begin{table*}[htbp]
\centering
\caption{Threshold consistency. All selected measures, except GradSim, maintained stable thresholds with larger image counts (Table \ref{tab:Study_Data_description}). Poor Image quality in the Traveling Human Phantom dataset led to an incorrect threshold estimate in PCC, denoted by ${\ast}$. }
\renewcommand{\arraystretch}{1.2}
\begin{tabularx}{\linewidth}{|l|l|l|l|l|X|}
\hline
\textbf{Purpose} &
  \textbf{Datasets} &
  \textbf{Threshold SSIM} &
  \textbf{Threshold NMI} &
  \textbf{Threshold PCC} &
  \textbf{Threshold GradSim} \\ \hline
\multirow{2}{*}{\textbf{Pre Evaluation}} & SLDM & 0.827 & 0.478 & 0.968 & 0.917 \\ \cline{2-6} 
                                         & SHCP & 0.851 & 0.510 & 0.980 & 0.934 \\ \hline
\multirow{5}{*}{\textbf{Evaluation}} &
  ADNI &
  0.885 &
  0.517 &
  0.987 &
  0.947 \\ \cline{2-6} 
                                         & Running intervention Dataset & 0.759 & 0.439 & 0.968 & 0.902 \\ \cline{2-6} 
                                         & Traveling Human Phantom & 0.791 & 0.441  & $0.781^{\ast}$ & 0.920 \\ \cline{2-6} 
                                         & Hormonal Health Study (HHS) & 0.739 & 0.416 & 0.961 & 0.895 \\ \cline{2-6} 
                                         & SDSU-TS & 0.717 & 0.390 & 0.961 & 0.897 \\ \hline
\end{tabularx}%

\label{tab:thresholding_values}
\end{table*}

\section{Discussion}
Previous research has shown that brain MRIs can be matched to other brain scans from the same person. 
Here, we asked whether the MRIs of the same person could be matched across different databases. We demonstrated that accurate participant matching is achievable with standard methods in an unsupervised manner in various simulated scenarios across databases. Notably, MRIs from different sites, scanner types, and time intervals could be matched even when with participants suffering from cognitive decline. Our MRI matching pipeline was particularly simple: we applied standard preprocessing to harmonize 3D T1-weighted scans and computed their pairwise similarity, detecting the scans from the same person by thresholding the similarity measures. We validated our approach on two simulated datasets and tested it extensively on five diverse public datasets covering longitudinal, multi-protocol, and cross-site scenarios. Matching accuracy was nearly perfect, with AUC reaching 1.000 in most cases.

 Previous research has studied MRI-based participant identification based on supervised learning, typically requiring a training set of at least two scans per participant: Valizadeh et al. \cite{valizadeh2018identification} demonstrated it using T1-weighted structural MRI, achieving near-perfect identification (F1 score = 0.90–0.99) with FreeSurfer-derived anatomical features. Finn et al.\cite{finn2015functional} reported 98–99\% identification accuracy using functional connectivity patterns from fMRI, with further improvements by Amico et al. \cite{amico2018quest}. More recently, DeepBrainPrint \cite{puglisi2023deepbrainprint} and Ogg et al. \cite{ogg2024large} reported identification accuracies in the 94–99\% range.
 In contrast to prior work, the approach presented here does not need paired training data but is fully unsupervised and relies only on computationally efficient methods. We showed that simple MRI preprocessing, when applied to skull-stripped brain images, was sufficient to expose implicit participant-specific features and reveal intra-participant links without training. Standard similarity measures such as PCC or SSIM were sufficient to match MRIs from the same participant. In addition, affine registration alone was sufficient, indicating that deformable transformations are not necessary for MRI matching. The keypoint‑based similarity measure of Chauvin et al.\cite{chauvin2020neuroimage} most closely resembles our matching framework. However, because their work did not address privacy, they did not evaluate same‑participant matching or cross‑database scenarios. Our proposed pipeline is also simpler than theirs.

Our results underscore the importance of continuous monitoring and assessment of developments in data-processing technologies in rapidly evolving fields such as ML and AI. Technological progress can easily affect the boundaries between anonymization and pseudonymization. This underlines the need for those controlling data not only to stay informed about technological developments but also to carefully consider their legal implications. The legal implications may vary depending on applicable law, its interpretations, and guidance given for it.

As an example of the legal assessment needed, we can take a data controller acting under the GDPR. The legal assessment made by such data controller culminates to two questions. First question to be answered is on the likelihood of means of re-identification: does there exist “means reasonably likely to be used to identify the natural person”, as stated in Recital 26 of the GDPR. Second question requiring answer is whether the residual risk level embedded in the possibility to link pseudonymized data to other information is “insignificant”, as addressed in the European Data Protection Board (EDPB) guidance.  To answer such questions, the data controller should investigate properly the specific context at hand and detailed flows of data processing activities relating thereto, evaluate possible risks embedded in quasi-identifiers, and assess relevant pseudonymization domains by adopting a relative perspective based on the recipient’s viewpoint.

The first question on the likelihood of means of re-identification is based on Recital 26 of the GDPR. The legal assessment of whether specific information can be considered as personal data depends on whether a natural person is identified or identifiable from the information. Pseudonymized personal data that could be attributed to a natural person by the use of additional information is considered identifiable. However, in this respect, only the means “reasonably likely” to be used should be taken into account. It should specifically be noted that unlawful means are not considered reasonable. Additionally, it should be kept in mind that all recipients are automatically restricted by the requirements set out in law, e.g., to the lawfulness of processing, purpose limitation, and data minimization. This stage of the assessment thus rules out the need to take into account for instance unlawful means when considering the re-identification challenges relating to pseudonymized data.

Regarding the second question on the level of residual risk, the role of quasi-identifiers, i.e. the combination of attributes making it possible to attribute pseudonymized data to data subjects without the use of pseudonymisation secrets like keys or matching tables, needs to be considered. Quasi-identifier is a legal construction and has been addressed by the EDPB in its Guidelines 01/2025 on Pseudonymisation, adopted on 16 January 2025\cite{EDPB2025Pseudonymisation}. According to the EDPB Guidelines, there are several ways to treat quasi-identifiers, such as removal, modification, or limiting access by technical controls. The EDPB Guidelines conclude that ``If an assessment shows that there is an insignificant risk that the pseudonymised data are linked to other information, then quasi-identifiers may be kept.”

When assessing both the means of re-identification and the residual risk level, the need to define the pseudonymisation domain, i.e., the context of pseudonymisation, becomes essential. This requires consideration of the data protection from a relative perspective, i.e., from the recipient’s perspective, instead of an absolute one. The risk should be assessed per each relevant pseudonymisation domain. In other words, the same data can be regarded as personal data for an original controller, pseudonymised data for further controllers having legal access to pseudonymised data, and anonymous for a further group of recipients. This approach is in line with the Court of Justice’s case law (Brayer Judgement C-582/14 from 2016\cite{CJEU_C-582_14_Breyer_2016}; Scania Judgment C-319/22 from 2023\cite{CJEU_C-319_22_Gesamtverband_AH_v_Scania_2023}; and EDPS v SRB Opinion from 2025\cite{CJEU_C-413_23P_EDPS_v_SRB_2025}).

As demonstrated in this article, T1w MRI can serve as a link between other MRIs of the same participant. For a brain MRI to enable re-identification, several additional steps would be required. These include the need to use tools (readily available as open-source), but additionally, acts combining research data (readily available as open data) with real-world data. The real-world data has several access restrictions, and presumably, most of the tasks combining research data to real-world data would require unlawful means, i.e., acting against the grounds for lawfulness of processing, purpose limitation, or data minimization. As unlawful means are not considered reasonable, the need to use real-world-data would cause the MRI to remain, in most cases, outside the notion of personal data. Despite regarding T1w MRI as non-identifiable to a person in the public sphere, the study highlights the need for safeguards to be taken against unlawful access to the real-world data in other than public environments. Special attention should be given to the organizational measures employed, as the possibility that MRI can serve as a link exists, given readily available open-source tools that enable re-identification. 

Organizational measures could involve data collection entities implementing clear governance policies that account for prior research participation, where feasible, and transparently informing participants when data may be shared or open-sourced. Explicit communication of potential downstream risks, including cross-study data linkage, should be integrated into the consent process (in case a participant has participated in multiple open-source studies), and DPIAs should be conducted prior to data release. These measures are significant given that re-identification risk is cumulative: while a single study may pose minimal risk, repeated participation across studies and modalities can facilitate linkage, especially in small or highly specific populations.

We showed that the MRI linkage property remained robust even with disease-related structural changes using the ADNI dataset. Although neurodegeneration associated with dementia lowered similarity scores, these values stayed above the identification threshold, indicating that individual-specific anatomical signatures persist despite pathology. 
This finding raises further questions about the features that link two similar images despite disease-related changes. Future studies should investigate the specific anatomical, morphometric, or representational features driving similarity to better understand how reliable matching occurs even with minimal preprocessing and structural pathology.

Future research should also evaluate how standard pre-processing affects the detection of original information in synthetic T1-weighted MRI data. Testing raw synthetic data without simulation could determine whether basic preprocessing and simple similarity measures are sufficient to validate synthetic T1-weighted brain MRI generated by deep learning methods.
In addition, other imaging modalities or anatomical regions beyond the brain should be investigated with respect to linking capability across databases.

To conclude, our findings confirm that T1w head MRI, post-skull-
stripping can still serve as a link between two datasets, making MRI a potential link between databases. 
These findings support the implementation of stronger privacy safeguards and responsible data governance.
We suggest that consent forms should include a question about prior participation in research projects including MRI scans. This additional context can help participants proactively inform data controllers if they have previously contributed MRI data. These results aim to contribute meaningfully to the development of thoughtful, forward-looking policies in medical data sharing.

\bibliographystyle{IEEEbib}
\bibliography{references}

\section{Acknowledgements}
This research, conducted as part of the PhaseIVAI project (133873) funded by the European Union, has received support from the EU’s Horizon Europe research and innovation programme under grant agreement No 101095384.

Additionally, it has been supported by grants 346934 and 358944 (Flagship of Advanced Mathematics for Sensing Imaging and Modeling) from the Research Council of Finland, as well as grant 351849 from the Research Council of Finland under the ERA PerMed framework (“Pattern-Cog”).

Data collection and sharing for this project was funded by the Alzheimer’s Disease Neuroimaging Initiative (ADNI) (National Institutes of Health Grant U01 AG024904) and DOD ADNI (Department of Defense award number W81XWH-12-2-0012). ADNI is funded by the National Institute on Aging, the National Institute of Biomedical Imaging and Bioengineering, and through generous contributions from the following: AbbVie, Alzheimer’s Association; Alzheimer’s Drug Discovery Foundation; Araclon Biotech; BioClinica, Inc.; Biogen; Bristol-Myers Squibb Company; CereSpir, Inc.; Cogstate; Eisai Inc.; Elan Pharmaceuticals, Inc.; Eli Lilly and Company; EuroImmun; F. Hoffmann-La Roche Ltd and its affiliated company Genentech, Inc.; Fujirebio; GE Healthcare; IXICO Ltd.; Janssen Alzheimer Immunotherapy Research \& Development, LLC.; Johnson \& Johnson Pharmaceutical Research \& Development LLC.; Lumosity; Lundbeck; Merck \& Co., Inc.; Meso Scale Diagnostics, LLC.; NeuroRx Research; Neurotrack Technologies; Novartis Pharmaceuticals Corporation; Pfizer Inc.; Piramal Imaging; Servier; Takeda Pharmaceutical Company; and Transition Therapeutics. The Canadian Institutes of Health Research is providing funds to support ADNI clinical sites in Canada. Private sector contributions are facilitated by the Foundation for the National Institutes of Health (\url{http://www.fnih.org}). The grantee organization is the Northern California Institute for Research and Education, and the study is coordinated by the Alzheimer’s Therapeutic Research Institute at the University of Southern California. ADNI data are disseminated by the Laboratory for Neuro Imaging at the University of Southern California.

Data were provided [in part] by the Human Connectome Project, WU-Minn Consortium (Principal Investigators: David Van Essen and Kamil Ugurbil; 1U54MH091657) funded by the 16 NIH Institutes and Centers that support the NIH Blueprint for Neuroscience Research; and by the McDonnell Center for Systems Neuroscience at Washington University.

The computational analyses were partly performed on servers provided by the UEF Bioinformatics Center and Biocenter Kuopio, Biocenter Finland, University of Eastern Finland, Finland.

\section{Competing interests}
All authors declare no financial or non-financial competing interests.

\section{Ethics and Consent to Participate}
This study used secondary datasets, and no additional ethics approval or consent was required.
All datasets analyzed in this study were previously collected by the original investigators with appropriate informed consent.

\section{Data Availability}

All datasets analyzed in this study were previously collected by the original investigators with appropriate informed consent, and the present work constitutes a secondary analysis of publicly available data. Specifically, we utilized data from the Alzheimer’s Disease Neuroimaging Initiative (ADNI), the OpenNeuro database (accession numbers: ds000206, ds004937, ds005664, and ds005360), the Human Connectome Project (HCP), and the LDM100K dataset.

The ADNI data used in the preparation of this article were obtained from the ADNI database (https://adni.loni.usc.edu). Investigators within ADNI contributed to the design and implementation of ADNI and/or provided data but did not participate in the analysis or writing of this report. A complete listing of ADNI investigators is available at: \url{http://adni.loni.usc.edu/wp-content/uploads/how_to_apply/ADNI_Acknowledgement_List.pdf}.

OpenNeuro is an open-access repository of de-identified neuroimaging datasets that are shared for secondary research use. The repository is publicly accessible at \url{https://openneuro.org/}.

The HCP Young Adult Open Access dataset, collected by the Washington University–University of Minnesota Consortium of the Human Connectome Project (WU-Minn HCP), was used in accordance with the HCP Data Use Terms. 

In addition, this study included the LDM100K dataset, a synthetically generated neuroimaging dataset. The synthetic data model which was used to generate the data was trained with data derived from UK Biobank by the original authors of the dataset.

All datasets used in this study are either publicly available and de-identified (OpenNeuro), synthetic (LDM 100K), or available to qualified researchers upon account registration and agreement to the data usage conditions (ADNI and HCP). They are properly cited in the article, and if the participants were randomly selected, their IDs are provided in the Supplementary Information.

\section{Code Availability}
The underlying code for this study is publicly available on URL: \url{https://extgit.vtt.fi/gaurang.sharma/mri_matching}

\section{Supplementary Information}
Yes, this manuscript includes supplementary information.

\section{Author Contribution}

G.S. developed the methods, conducted experiments, analyzed data, and drafted the first version of the manuscript. HP contributed to methods and data analysis. J.S. provided feedback on data governance, compliance, and interpretation. J.P. conceived the idea for the study and contributed to the methods and data analysis. J.T. supervised the study and contributed to methods and data analysis. All authors reviewed, revised for intellectual content, and approved the final manuscript.

\section*{Appendix}
\label{Appendix}

Similarity between the harmonized images was assessed voxel-wise. We implemented 11 similarity measures, most using custom code based on \texttt{torch}, \texttt{cupy} (for GPU acceleration), \texttt{numpy}, and \texttt{skimage}, while a subset relied on existing PyTorch-based implementations. The evaluated measures are listed below. Below are the details of each measure, where for each image pair \((X, Y)\), we denote the voxel intensities at location \(i\) by \(X_i\) and \(Y_i\). Also, we used a small constant \(C = 10^{-6}\) to ensure numerical stability, and represented the total number of voxels by \(N\).

\subsubsection{Mutual Information (MI)}
MI measures how much the intensity distribution of one image decreases the entropy of the other. Unlike correlation, MI captures nonlinear relationships, making it effective for comparing images from different modalities. 
MI was computed as:
\[
\mathrm{MI} = \sum_{x}\sum_{y} p(x,y)\,\log\!\left(\frac{p(x,y)}{p(x)\,p(y)}\right),
\]
where \(p(x,y)\) is the joint probability of intensity \(x\) in \(X\) and intensity \(y\) in \(Y\), and \(p(x)\) and \(p(y)\) are the corresponding marginal probabilities.
The joint probability distribution $p(x,y)$ was estimated using a two-dimensional histogram of paired voxel intensities with 50 bins. Histogram counts were normalized by the total number of voxel pairs to obtain $p(x,y)$. The marginal distributions $p(x)$ and $p(y)$ were then computed by summing the joint distribution over the corresponding dimensions:
\[
p(x_i) = \sum_j p(x_i,y_j), \qquad
p(y_j) = \sum_i p(x_i,y_j).
\]

\subsubsection{Normalized Mutual Information (NMI)}
NMI measures the shared information between two
images by comparing their joint and marginal entropy. It is robust to intensity
scaling differences and is widely used in multimodal image registration. 
The measure was computed as
\[
\mathrm{NMI} = \frac{\mathrm{MI}}{\sqrt{H(X)\,H(Y)}},
\]
where \(\mathrm{MI}\) is the mutual information, and the marginal entropies are
defined as
\[
H(X) = -\sum_{x} p(x)\,\log p(x) \]

\[
H(Y) = -\sum_{y} p(y)\,\log p(y).
\]

\subsubsection{Negative Mean Squared Error (NMSE)}
NMSE calculates the average squared difference between
corresponding voxels, multiplied by $-1$. It is simple and widely used, but sensitive to outliers
and does not always reflect perceptual similarity. 
The measure was computed as
\[
\mathrm{NMSE} = - \frac{1}{N}\sum_{i}(X_i - Y_i)^2.
\]

\subsubsection{Peak Signal-to-Noise Ratio
(PSNR)}
PSNR is derived from MSE, and it measures the ratio between the maximum possible signal intensity
and the distortion introduced. 
The measure was computed as
\[
\mathrm{PSNR} = 10\,\log_{10}\!\left(\frac{L^2}{\mathrm{MSE}}\right),
\]
where \(L\) is the maximum possible voxel intensity value, and MSE is the
mean squared error between the images (-NMSE).

\subsubsection{Pearson’s Correlation Coefficient
(PCC)}
PCC measures the linear correlation between
two images by comparing the covariation of their voxel intensity values. It
ranges from $-1$ to $+1$, where $+1$ indicates perfect linear similarity. This
measure is useful when the relative pattern of intensities matters more than
absolute differences. The measure was computed as
\[
\mathrm{PCC} =
\frac{\sum_i (X_i - \bar{X})(Y_i - \bar{Y})}
     {\sqrt{\sum_i (X_i - \bar{X})^2}\,\sqrt{\sum_i (Y_i - \bar{Y})^2}},
\]
where 
$\bar{X}$ and $\bar{Y}$ are the
average voxel intensities of the respective images.

\subsubsection{Cosine Similarity (CosSim)}
CosSim measures the cosine of the angle between two flattened image vectors. It
evaluates whether the images point in the same direction in a high-dimensional space
, independent of their magnitude. Values range from $-1$ to $+1$, with
$+1$ indicating identical patterns up to a scale factor. 

We used the open-source \texttt{torch.nn.functional} module to compute the measure as:
\[
\mathrm{CosSim}
= \frac{\sum_i X_i Y_i}
       {\sqrt{\sum_i X_i^2}\,\sqrt{\sum_i Y_i^2}}.
\]
The numerator measures how strongly the two vectors align, while
the denominator normalizes by their magnitudes, so the resulting value reflects
only the similarity of their directional patterns.

\subsubsection{Gradient Similarity (GradSim)}
GradSim\cite{6081939} compares the spatial gradient magnitudes (edge strength) of two images rather than their raw intensity values. It evaluates the preservation of edges, contours, and intensity transitions associated with image structure.
This measure between two 3D images $X$ and $Y$ was computed voxel-wise as
\begin{equation}
\mathrm{GradSim} =
\frac{
2\,\lvert \nabla X_\mathbf{i} \rvert \, \lvert \nabla Y_\mathbf{i} \rvert + C
}{
\lvert \nabla X_\mathbf{i} \rvert^{2} + \lvert \nabla Y_\mathbf{i} \rvert^{2} + C
},
\end{equation}
where $\nabla X$ and $\nabla Y$ denote the image gradients computed using
3D Sobel kernels and $\lvert \nabla X \rvert$ and $\lvert \nabla Y \rvert$ represent
the corresponding gradient magnitudes (edge strength).

The final GradSim score between $X$ and $Y$ is obtained by spatially averaging
the voxel-wise similarity over the image domain $\Omega$:
\begin{equation}
\mathrm{GradSim}(X,Y)
=
\frac{1}{\lvert \Omega \rvert}
\sum_{\mathbf{i} \in \Omega}
\mathrm{GradSim}.
\end{equation}

\subsubsection{Structural Similarity Index
(SSIM)}
SSIM\cite{1284395} assesses image similarity using three components:
luminance, contrast, and structure.
Values range from $-1$ to $1$, with $1$
indicating perfect structural similarity. The measure was computed as
\[
\mathrm{SSIM} =
\frac{(2\mu_X\mu_Y + C_1)(2\sigma_{XY} + C_2)}
     {(\mu_X^2 + \mu_Y^2 + C_1)(\sigma_X^2 + \sigma_Y^2 + C_2)},
\]
where $\mu_X$ and $\mu_Y$ are the mean intensities, $\sigma_X^2$ and $\sigma_Y^2$
are the variances, $\sigma_{XY}$ is the covariance.

We computed SSIM using a PyTorch-based \texttt{torchmetrics.functional} implementation with default parameters, including a data range of $255$, spatial averaging, and a Gaussian weighting
window of size $11$ with a standard deviation $1.5$. It assumed $3$
input channels and two spatial dimensions, with stability constants
$(K_1, K_2) = (0.01, 0.03)$. Non-negative SSIM enforcement was not applied.

\subsubsection{Multi-Scale Structural Similarity (MS-SSIM)}
MS-SSIM\cite{wang2003multiscale} extends SSIM by evaluating similarity across multiple
image resolutions. 
The measure was computed as
\[
\mathrm{MS\text{-}SSIM} = \prod_{j}\bigl[\, l_j^{\alpha_j}\, c_j^{\beta_j}\, s_j^{\gamma_j}\,\bigr],
\]
where the luminance, contrast, and structure comparisons at scale \(j\) are given by
\[
l_j = \frac{2\mu_{X_j}\mu_{Y_j} + C_1}{\mu_{X_j}^2 + \mu_{Y_j}^2 + C_1},\qquad
c_j = \frac{2\sigma_{X_j}\sigma_{Y_j} + C_2}{\sigma_{X_j}^2 + \sigma_{Y_j}^2 + C_2},
\]
\[
s_j = \frac{\sigma_{X Y_j} + C_3}{\sigma_{X_j}\sigma_{Y_j} + C_3},
\]
where $j$ denotes the scale level (e.g., coarse to fine), $l_j$, $c_j$, and $s_j$ correspond respectively to luminance, contrast, and structure comparisons, $\alpha_j$, $\beta_j$, and $\gamma_j$ are the weights for each scale, $\mu_{X_j}$ and $\mu_{Y_j}$ denote the local mean intensities of the image patches, $\sigma_{X_j}$ and $\sigma_{Y_j}$ denote the corresponding local standard deviations, and $\sigma_{XY_j}$ denotes the local cross-covariance between the two image patches.

MS-SSIM was computed using an open-source PyTorch \texttt{pytorch\_msssim} implementation with default settings, matching the SSIM configuration. The measure employed the standard multi-scale formulation with default scale weights.

\subsubsection{Four-Component Gradient-Regularized SSIM (4-G-R-SSIM)}
Four-Component Gradient-Regularized SSIM (4-G-R-SSIM) \cite{renieblas2017structural} extends the standard SSIM by evaluating structural similarity on directional image gradients, thereby increasing sensitivity to edge information and structural distortions. The method computes SSIM on gradient images in four orientations and combines the resulting similarity maps into a rotation-aware measure. 

For each directional gradient $d \in \{1,\dots,4\}$, local Gaussian-weighted statistics $\mu_X^{(d)}$, $\mu_Y^{(d)}$, $\sigma_X^{2(d)}$, $\sigma_Y^{2(d)}$, and $\sigma_{XY}^{(d)}$ were computed from the gradient images, and the gradient SSIM was defined as
\[
\mathrm{GSSIM}^{(d)} = \frac{(2\mu_X^{(d)}\mu_Y^{(d)} + C_1)(2\sigma_{XY}^{(d)} + C_2)}{(\mu_X^{(d)2} + \mu_Y^{(d)2} + C_1)(\sigma_X^{2(d)} + \sigma_Y^{2(d)} + C_2)},
\]
where $C_1=(k_1L)^2$ and $C_2=(k_2L)^2$. The four directional responses were fused by averaging, and the final 4-G-R-SSIM score was obtained by averaging over all voxels.

In our implementation, axial slices were processed independently. For each slice, gradients were computed in four directions, local gradient-based SSIM was evaluated, and the fused similarity maps were averaged across all slices and voxels to yield the final 3D similarity score.

\subsubsection{Negative Fr\'echet Inception Distance
(NFID)}
NFID\cite{NIPS2017_8a1d6947} measures the similarity between two distributions
of images (e.g., real vs.\ generated) using deep feature embeddings. It computes
the distance between their multivariate Gaussian statistics (means and
covariances). However, for comparing pairs of 3D MRI volumes, we used an
MRI-adapted version of NFID. Unlike the original formulation, which compares
distributions of many images using features from an Inception network, here each
MRI scan is treated as a distribution over voxels.

Each MRI \(X\) and \(Y\) was first resampled to a fixed resolution of
\(96\times 96\times 96\) and intensity normalized. A 3D DenseNet\textendash121
encoder (\(\text{spatial\_dims}=3\), \(\text{in\_channels}=1\),
\(\text{out\_channels}=1024\)) was then used to extract feature embeddings from
each voxel. 

The measure was computed as
\[
\mathrm{NFID} = -\lVert \mu_X - \mu_Y \rVert^2
+ \operatorname{Tr}\!\left( \Sigma_X + \Sigma_Y
- 2\,(\Sigma_X\,\Sigma_Y)^{1/2} \right),
\]
where \(\mu_X\) and \(\mu_Y\) are the mean feature vectors of images \(X\) and
\(Y\), \(\Sigma_X\) and \(\Sigma_Y\) are the corresponding covariance matrices,
\(\lVert \mu_X - \mu_Y \rVert^2\) denotes the squared Euclidean distance, and
\(\operatorname{Tr}(\cdot)\) is the matrix trace, capturing differences in the
covariance structure of the two distributions being compared.

\end{document}